\documentclass[10pt,twocolumn,letterpaper]{article}   
\usepackage[pagenumbers]{cvpr}

\usepackage{graphicx}
\usepackage{amsmath}
\usepackage{amssymb}
\usepackage{booktabs}
\usepackage{array}
\usepackage{color}
\usepackage{makecell}
\usepackage[dvipsnames]{xcolor}

\usepackage{lipsum}
\usepackage[normalem]{ulem}

\usepackage{pifont}

\usepackage{setspace}
\usepackage{graphicx}

\usepackage{colortbl}
\usepackage[dvipsnames]{xcolor}

\usepackage{enumitem}
\usepackage[pagebackref,breaklinks,colorlinks,bookmarks=false]{hyperref}

\usepackage[capitalize]{cleveref}
\crefname{section}{Sec.}{Secs.}
\Crefname{section}{Section}{Sections}
\Crefname{table}{Table}{Tables}
\crefname{table}{Tab.}{Tabs.}

\def\cvprPaperID{32} 
\def\confName{CVPRW}
\def\confYear{2022}

\begin{document}

\title{Negative Frames Matter in Egocentric Visual Query 2D Localization}

\author{Mengmeng Xu$^{1,2}$
\and
Cheng-Yang Fu$^{1}$
\and
Yanghao Li$^{1}$
\and
Bernard Ghanem$^{2}$
\and \newline
Juan-Manuel P\'erez-R\'ua$^{1}$
\and
Tao Xiang$^{1,3}$ 
\and \newline
{\small $^1$ Meta AI, UK} ~ 
{\small $^2$ King Abdullah University of Science and Technology, Saudi Arabia} ~
{\small $^3$ University of Surrey, UK} \\
}

\maketitle

\begin{abstract}
The recently released Ego4D dataset and benchmark~\cite{Ego4D2022CVPR} significantly scales and diversifies the first-person visual perception data. 
In Ego4D, the Visual Queries 2D Localization task aims to retrieve objects appeared in the past from the recording in the first-person view. This task requires a system to spatially and temporally localize the most recent appearance of a given object query,
where query is registered by a single tight visual crop of the object in a different scene. 

Our study is based on the three-stage baseline introduced in the Episodic Memory benchmark. The baseline solves the problem by detection and tracking: detect the similar objects in all the frames, then run a tracker from the most confident detection result. 
In the VQ2D challenge, we identified two limitations of the current baseline. 
(1) The training configuration has redundant computation. Although the training set has millions of instances, most of them are repetitive and the number of unique object is only around 14.6k. The repeated gradient computation of the same object lead to an inefficient training;  
(2) The false positive rate is high on background frames. This is due to the distribution gap between training and evaluation. During training, the model is only able to see the clean, stable, and labeled frames, but the egocentric videos also have noisy, blurry, or unlabeled background frames.
To this end, we developed a more efficient and effective solution. Concretely, we bring the training loop from ~15 days to less than 24 hours, and we achieve $0.17\%$ spatial-temporal AP, which is $31\%$ higher than the baseline. Our solution got the first ranking on the public leaderboard. Our code is publicly available at \url{https://github.com/facebookresearch/vq2d_cvpr}.
\end{abstract}

\section{Introduction}
The task of Visual Queries 2D Localization in egocentric videos aims to retrieve objects from episodic memory. This task can be described as `when was the last time that I saw X', where X is a query represented by a visual crop. In a real-world application, X can be the user's wallet -- after registering the wallet image as an image crop, the user can easily retrieve the wallet's location by a visual query system. 

As a formal definition, given a visual crop $v$ of a query object $o$ and a query frame (time) $q$, the system should provide a `response track' $r$ that follows $o$  before frame $q$. The response track $r$ is a temporally contiguous set of bounding boxes surrounding the object $o$. $r = \{r_s,r_{s+1},\cdots ,r_{e-1},r_e\}$, where $s$ and $e$ are the first and last frames that the object is visible, respectively, and $r_i$ represents a bounding box $(x, y, w, h)$ in frame $i$.

The predictions are mainly evaluated by spatio-temporal Average-Precision (stAP)~\cite{gkioxari2015finding}. It measures how closely the predicted bounding box set matches with the ground-truth response track in the spatio-temporal extent. 

\begin{figure}\centering
\includegraphics[trim={8cm 3.5cm 8cm 2cm},width=8.5cm,clip]{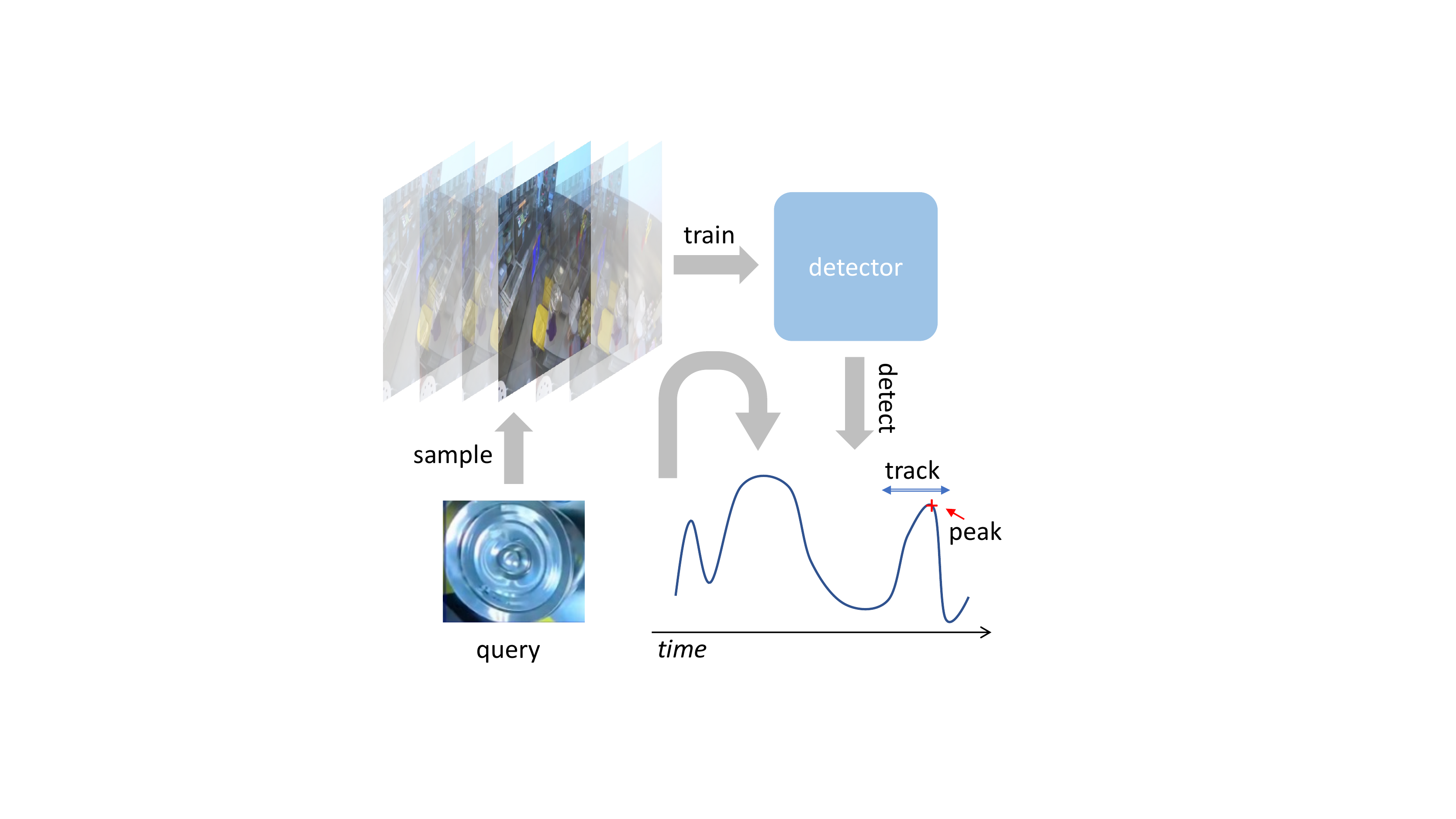}
\caption{\textbf{Baseline of the visual queries 2D localization task.} Given a trained detector, the response track are generated via three steps. (1) Detect the query object in all the frames; (2) localize the most confident frame that includes the query object (3) track the spatio-temporal localized object on two direction to generate the response track.
}
\label{fig:baseline}
\end{figure}

The VQ2D baseline is a three-step detection + tracking solution. We summarize the process in Fig.~\ref{fig:baseline}. Firstly, an object detector is trained using frame-crop pairs. The model inputs are a query object crop and a video frame, and the output is the object in the frame that matches the crop. When the training of the detection model is converged, we apply the detector to all the frames in the video, then create a similarity curve, where each point scores the similarity of the top-1 prediction in the frame to the query object. From the curve we can detect the peak and locate the bounding box in the peak frame. Finally, the response object track is generated by running a tracker in forward and backward directions from the peak bounding box. 

Our solution is developed from the baseline with the following improvements: 

(1) Accelerated training and evaluation. The original baseline requires around 3600 hours to finish the training loops for the object detection model in step 1, and it takes nearly 400 hours for a full evaluation in step 2 and 3. We explore a new set of training parameters aiming at training efficiency. Additionally, we refactor training and evaluation code to enable parallel computing. This results in optimized training and evaluation process, reducing execution time to 24 hours and 12 hours, respectively. This means we have improved the speed of the research time cycle to $1500\%$ in training and $3200\%$ in testing. The improvement significantly shortens the experiment loop and can give fast feedback to new research ideas for us and the community.

(2) Error type analysis. We also study the error in detection and tracking results. The model produces false positives on negative frames, which is not reflected in the default detection evaluation. This is because the detection evaluation metric in step 1 disregards background frames and leads to a discrepancy in the final retrieval evaluation (stAP). The detection metric only requires the model to have a high recall but this VQ2D task also requires a high-precision system. We visualize different types of false-positive errors in negative frames in the video and show that simply training our model with extra negative frames can bring a performance gain in the final evaluation. We wish this exposition to the community can lead to more investigations in this area.

(3) A more precise implementation. Our model can surpass the state-of-the-art by $0.02$ in stAP25, by $3.8\%$ in success rate, and by $4.1\%$ in recovery ratio on the validation set. On the test set, we achieve $0.17\%$ spatial-temporal AP, which is $31\%$ higher than the baseline. The main reason for such large improvement is that we optimize the backbone detector with egocentric data: we identified that the pre-trained model can increase its capability to represent Ego4D data by backbone optimization, and training with negative frames which do not include query objects can reduce false positives of the detector. 


\begin{figure*}[h]
\centering
\includegraphics[trim={3.5cm 2cm 3.5cm 0cm},width=17cm,clip]{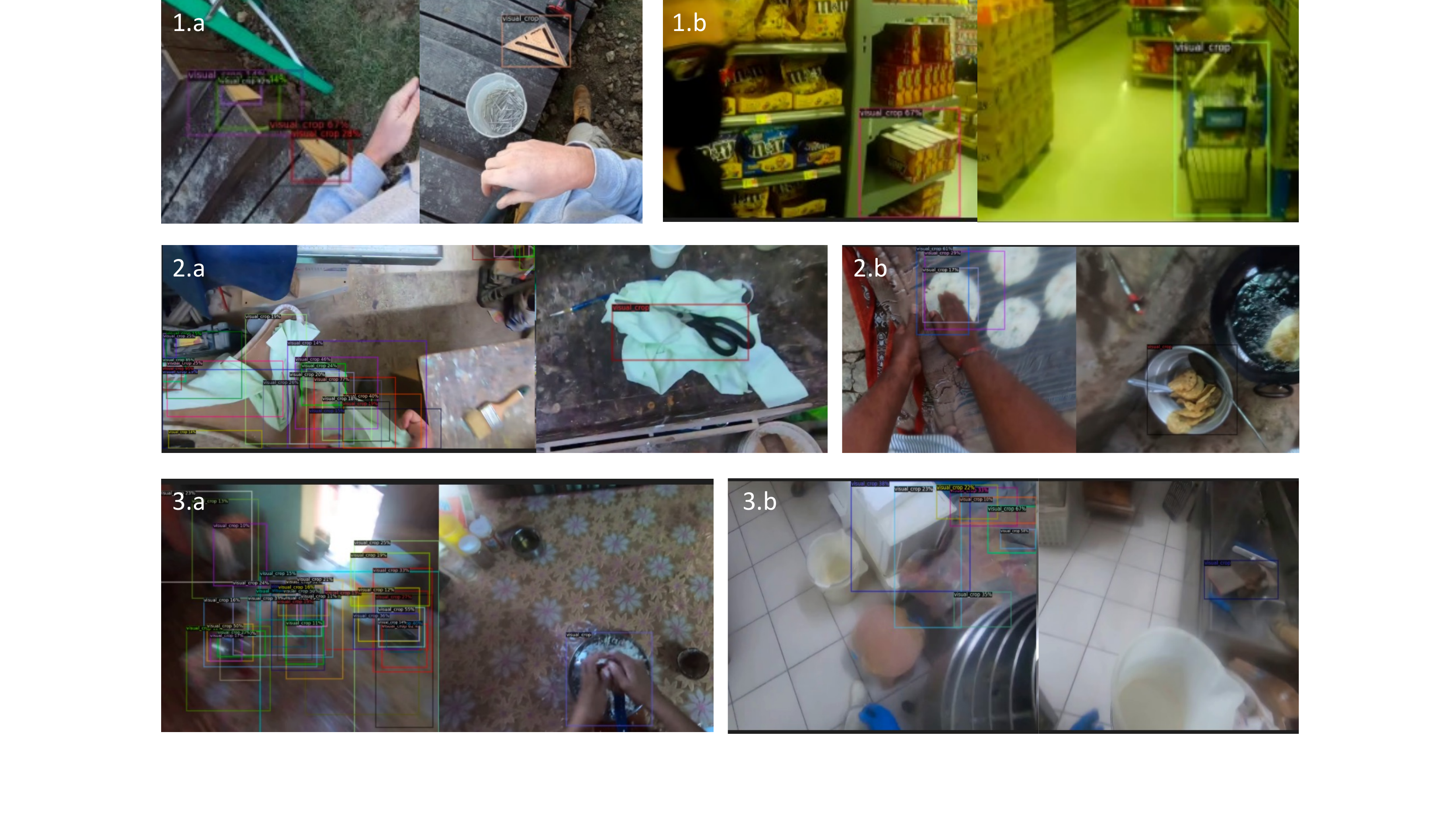}
\caption{\textbf{Visualization of false positives in negative frames.} In each example, the bounding boxes in the left image are from the detector, and the bounding box in the right image shows the visual crop. We categorize them into three error classes: similar appearance, unambiguous query, and blurry frame. 
}
\label{fig:vis}
\end{figure*}

\section{Faster Training and Evaluation}
The baseline object detector has three modules: visual backbone, region proposal network, and siamese region-of-interest (roi) head. 
Concretely, the visual backbone transforms both the input frame and the object crop into feature space. The region proposal network generates object bounding box candidates based on the frame feature. Then the siamese roi head compares every proposal with the object crop in the feature space and classifies all the proposals into positives and negatives. 
During training, both the visual backbone and the region proposal network are frozen, and the default 1 million training iterations carried by the baseline model are originally intended to optimize the siamese roi head alone. We identify two ways to improve the training process. Firstly, although the frozen visual backbone is pre-trained on COCO, a large-scale object detection dataset, we found that enabling backbone optimization can further reduce the domain gap between common objects and egocentric objects. Second, we discovered a more efficient learning rate schedule. Specifically, we reduce the learning rate at 20k, 50k, 100k, and 200k iterations in each individual experiment, as shown in Tab.~\ref{tab:iter} In the end, we have a 250k-iter setting that can achieve slightly better performance than the baseline, and a 125k-iter setting that can achieve the same performance as the baseline, but only 13 hours on a four-node machine. 

\begin{table}[ht]
    \centering
    \begin{tabular}{c|c|c|c|c|c}
    \toprule
     total iter & reduce lr & AP & AP50 & AP75 & AR@10  \\
      \midrule 
50k & 20k & 21.08 &	41.28	 & 17.82 & \textbf{44.6} \\
125k &50k & 23.26 &	45.00	 & \textbf{19.92} & 43.5 \\
250k &100k & \textbf{23.28} &	\textbf{45.31}	 & 19.90 & 43.9 \\
500k &200k & 22.51 &	44.68	 & 18.24 & 42.3 \\
    \bottomrule
    \end{tabular}
    \caption{We reduce the learning rate at different iterations (second column) in each individual experiment, then evaluate the model checkpoint after the last training iteration (first column). Due to the high training efficiency, we use the 125k-iter setting.}
    \label{tab:iter}
\end{table}

During evaluation, every possible visual query for video sequence is processed sequentially.
For each query, the detector is applied to all the frames in the video and the highest scoring proposal in that frame is recorded. The scores along the different timestamps of the video form a temporal curve, representing the confidence that the query object exists in the current frame. We re-implement the process to randomly group the videos and evaluate them in parallel. We create a parallel pool of size 100, reducing the total evaluation time to around 12 hours.

\section{Study on Failure Cases}\label{sec:neg}

We analyze the failure cases to understand the limitation of the system. We find the model presents a good recall, not suffering much from false negatives -- there are not many missing detected objects in the validation set. However, the false-positive rate is generally high. In particular, on background frames that do not have actual instances of the query object. One possible reason is that the model never sees such negative samples in both training and evaluation. In the visual query task, a high false-alarm rate will lead to false peaks in the temporal curve, leading to a completely wrong response track. We inspect the nature of these false positives and categorize them into three classes: similar appearance, unambiguous query, and blurry frame.

We give two examples in each row of Fig.~\ref{fig:vis}. In each example, the bounding boxes in the left image are from the detector, and the bounding box in the right image shows the visual crop.

Similar appearance error happens when the detector matches two objects that have a similar visual appearance but they have different semantics. In 1.a, the detector mismatches a triangle ruler as wooden stairs because they have similar colors and shapes. In 1.b, the detector misclassifies the boxes set on the shelf as a cart, and they have similar grid-like textures.

Ambiguous query error happens when the query object is absent in the frame. For example, the query object in 2.a is a scissor on a light-blue tower. When the scissor is missing in a negative frame, the detector will take the towel as positives by mistake. Similarly, in 2.b where the query object is the bucket, the detector learns to find the bread in the bucket when the bucket itself is missing.

The last row also showcases that when the frame is blurry, the detector tends to report more false positives. This is because we only use the clear image for training in the default setting, and the detector failed to interpret blurry images.

In summary, the detector is suffering from false positives, especially when the sampled frame does not have any positive object. Adding negative frames (Sec.~\ref{sec:train}) in training alleviates this issue, however, more technical work can be further explored to better tackle this problem.

\section{Training Detector with Negative Frames }\label{sec:train}

There are mainly two domain gaps between the detector training and the visual query task. The first domain gap is that the daily object in egocentric videos has a wider distribution than the COCO dataset used in pre-training. As a comparison, there are more than one thousand object classes in Ego4D, while COCO has only 80 classes of common objects. By enabling visual backbone optimization, our detector has the capability to learn from egocentric data to some extent, as shown in Tab.~\ref{tab:backbone}. As discussed in Sec.~\ref{sec:neg}, the second distribution gap is due to the missing negative frames in training. Especially when the query image is ambiguous or the frame image is with motion blur or out of focus. The baseline method already incorporates the idea of cross-batch negative sampling to discover negative objects from positive frames.  We further extend this concept by adding negative objects from background frames. Our motivation is to train the detector with both positive and negative frames such that it can be more robust when applied to all the frames in the evaluation process. Therefore, our proposal $p_j$ is negative if it satisfies any of the following two conditions: 
\begin{enumerate}
\item $j \in (s,e)$ and $IoU(p_j,r_j) < 0.5 $
\item $j \not\in (s,e) $
\item $p_j$ is sampled from another video. 
\end{enumerate}

Note that we also use hard-negative mining to select the top-K negatives with the highest loss value. We balance the positive-negative ratio to be $1:64$. See our result in Tab.~\ref{tab:nega}. Training with negative frames is able to reduce false positives and gives a better overall query response. We also found that the detection performance (AP) drops because this evaluation is only conducted on positive frames. When we evaluate on both positive and negative frames, the performance has an evident improvement from $25.95$ to $31.98$ in AP.

\begin{table}[ht]
    \centering
    \begin{tabular}{c|c|c|c|c|c}
    \toprule
      head lr & backbone lr & tAP25 & stAP25 & rec\% & Succ  \\
      \midrule 
    0.02 & frozen & 0.17 &	0.10 & 28.97 & 35.06 \\
    0.02 & 0.02 &	0.19 & 0.11  & 30.62 & 37.59 \\
    0.02 & 0.002 & \textbf{0.21} &	\textbf{0.14} & \textbf{34.37} & \textbf{42.21} \\
    \bottomrule
    \end{tabular}
    \caption{Optimizing visual backbone can reduce the domain gap between common objects and egocentric object.}
    \label{tab:backbone}
\end{table}

\begin{table}[ht]
    \centering
    \begin{tabular}{c|c|c|c|c}
    \toprule
      training & \multicolumn{2}{c|}{AP per frame} & \multicolumn{2}{c}{AP per video}   \\
      frames & pos. only & pos. + neg. & tAP25 & stAP25    \\
      \midrule 
    pos. only & \textbf{26.99} & 25.95 & {0.21} &	{0.14}  \\
    pos.+neg. & 26.28 & \textbf{31.98} & \textbf{0.22}	 & \textbf{0.15}	\\
    \bottomrule
    \end{tabular}
    \caption{Training with negative frames can reduce false positives and gives better query response. Note that the frame-level detection performance (AP) on the second column drops because it is only on positive frames, but the improvement is evident when evaluated with both positive and negative frames (3rd column).}
    \label{tab:nega}
\end{table}

\section{Conclusion}
In this report, we summarized the details of our solution to the Ego4D Visual Queries 2D Localization challenge. 
We discovered that improving training/evaluation speed is an essential step toward this task.
Also, we focus on alleviating the false positive issue which pays off in the final score. 
This results in achieving a top-1 performance on the public leaderboard.

\appendix
\vspace{5mm}
\noindent
\textbf{Participated Challenge}: Visual Queries 2D Localization

\noindent
\textbf{Participants}: 
\begin{itemize}[nosep]
\small
    \item Mengmeng Xu, KAUST, mengmeng.xu@kaust.edu.sa
    \item Juan-Manuel Pérez-Rúa, Facebook, jmpr@fb.com
    \item Cheng-Yang Fu, Facebook, chengyangfu@fb.com
    \item Yanghao Li, Facebook, lyttonhao@fb.com
    \item Bernard Ghanem, KAUST, bernard.ghanem@kaust.edu.sa
    \item Tao Xiang, Facebook, txiang@fb.com
\end{itemize}

\noindent \textbf{Acknowledgement}: 
The authors thank Kristen Grauman for discussion and Santhosh Kumar Ramakrishnan for his valuable help with baseline implementation. 

{\small
\bibliographystyle{ieee_fullname}
\bibliography{egbib}

\begin{thebibliography}{1}\itemsep=-1pt

\bibitem{gkioxari2015finding}
Georgia Gkioxari and Jitendra Malik.
\newblock Finding action tubes.
\newblock In {\em Proceedings of the IEEE conference on computer vision and
  pattern recognition}, pages 759--768, 2015.

\bibitem{Ego4D2022CVPR}
\href{https://sites.google.com/view/ego4d/home}{Ego4D Consortium 2020}.
\newblock Ego4d: Around the {W}orld in 3,000 {H}ours of {E}gocentric {V}ideo.
\newblock In {\em IEEE/CVF Computer Vision and Pattern Recognition (CVPR)},
  2022.

\end{thebibliography}
}

\end{document}